\documentclass[letterpaper,10pt,conference]{ieeeconf}
\IEEEoverridecommandlockouts
\usepackage{graphicx} 
\graphicspath{{./figures}}
\usepackage{amsmath,amsfonts,amssymb}
\usepackage{mathtools}
\usepackage{cite}
\usepackage{url}
\usepackage{booktabs}
\usepackage[dvipsnames]{xcolor}
\usepackage{multirow}
\usepackage[T1]{fontenc} 
\usepackage[caption=false,font=footnotesize]{subfig}

\title{\LARGE \bf Accurate Cooperative Localization Utilizing LiDAR-equipped Roadside Infrastructure for Autonomous Driving}

\author{Yuze Jiang$^1$, Ehsan Javanmardi$^1$, Manabu Tsukada$^1$, and Hiroshi Esaki$^1$
\thanks{$^1$Graduate School of Information Science and Technology, The University of Tokyo, Tokyo, 113-0033, Japan. (E-mail: \{uiryuu,ejavanmardi,mtsukada\}@g.ecc.u-tokyo.ac.jp, hiroshi@wide.ad.jp)}}

\begin{document}
\maketitle

\begin{abstract}
Recent advancements in LiDAR technology have significantly lowered costs and improved both its precision and resolution, thereby solidifying its role as a critical component in autonomous vehicle localization. Using sophisticated 3D registration algorithms, LiDAR now facilitates vehicle localization with centimeter-level accuracy. However, these high-precision techniques often face reliability challenges in environments devoid of identifiable map features. To address this limitation, we propose a novel approach that utilizes road side units (RSU) with vehicle-to-infrastructure (V2I) communications to assist vehicle self-localization. By using RSUs as stationary reference points and processing real-time LiDAR data, our method enhances localization accuracy through a cooperative localization framework. By placing RSUs in critical areas, our proposed method can improve the reliability and precision of vehicle localization when the traditional vehicle self-localization technique falls short. Evaluation results in an end-to-end autonomous driving simulator AWSIM show that the proposed method can improve localization accuracy by up to 80\% under vulnerable environments compared to traditional localization methods. Additionally, our method also demonstrates robust resistance to network delays and packet loss in heterogeneous network environments.
\end{abstract}

\section{Introduction}
Autonomous driving (AD) has advanced significantly over the past decade, propelled by developments in artificial intelligence and sensor technology. These innovations address various challenges as we progress toward higher levels of AD autonomy. Localization, a critical cornerstone of AD, is crucial because many safety-related AD services depend on the precise localization of the autonomous vehicle (AV)~\cite{Kuutti2018-rs}. Although vehicle self-localization techniques have rapidly improved in accuracy recently, their robustness remains unreliable in specific environments. 

On the other hand, it has become apparent that sensor occlusions and blind spots are critical issues for robust AD~\cite{Thakurdesai2021-ek}. Recently, to overcome these issues, infrastructure-aided cooperative perception has attracted significant research interest. This approach is favored not only for its greater computational power but also because sensors mounted on RSUs can provide a bird's-eye view (BEV), thereby efficiently eliminating blind spots across the entire vehicle-to-everything (V2X) network\cite{Cui2022-tj}.

\begin{figure}[t]
\centering
\includegraphics[width=\linewidth]{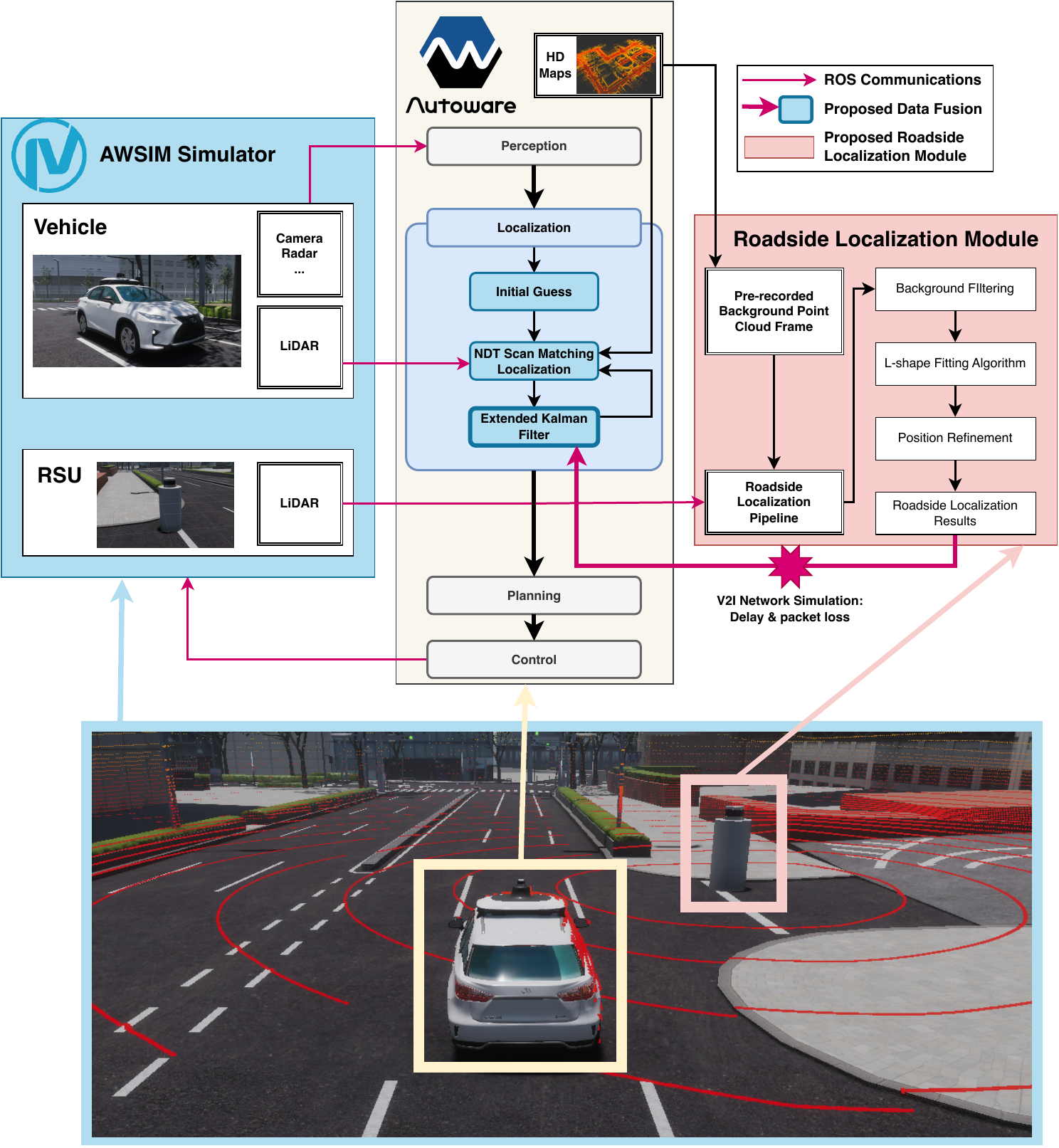}
\caption{The structure of the proposed system and experimental setup. Our proposed method involves the implementation of a roadside localization module, which includes conventional vehicle position estimation using roadside LiDAR and a position refinement algorithm, detailed in Section~\ref{sec:improvement}. To evaluate the performance of the proposed method, we conducted experiments in a virtual Nishishinjuku environment using the AWSIM simulator.}
\label{fig:structure}
\end{figure}

Therefore, we want to let vehicle localization also benefit from smart infrastructure. In this paper, we propose a cooperative localization service leveraging LiDAR-equipped infrastructure with dual objectives: 1) to provide highly accurate vehicle localization data, which can be fused with on-board data to enhance self-localization accuracy of CAVs, and 2) to augment the utility of existing sensing infrastructure, thereby increasing the value of RSUs. In our proposed method, the smart infrastructure acquires the dimensions of the connected autonomous vehicle (CAV) via V2I communications. This data is crucial for refining the accuracy of roadside perception by reducing position error. Subsequently, the RSU calculates the estimated 2D pose of the CAV based on their relative positions, and this information is sent back to the CAV. The received data is then fused with the localization results from the CAV’s on-board sensors, thereby improving its overall localization accuracy. Figure~\ref{fig:structure} illustrates the structure of our proposed method, the implementation of the roadside localization module, and the experiment setup.

In this work, we:
\begin{itemize}
    \item Proposed a highly-accurate cooperative localization method which uses roadside LiDAR.
    \item Conducted simulation-based experiments to evaluate the performance of the proposed method, during which data fusion with normal distribution transformation (NDT)-based localization occurred within a real AD stack.
    \item Demonstrated that our method can improve both the accuracy and stability of localization within the coverage area of the roadside LiDAR. We achieved sub 5 centimeters localization error when using VLP-32C inside the cooperative localization range.
\end{itemize}

\section{Related Work}
\subsection{High-Accuracy Roadside Perception}
Roadside perception is primarily utilized for object tracking, motion prediction, and speed validation~\cite{Zhao2019-jq}. Therefore, the absolute error of vehicle pose is often not evaluated in recent studies related to roadside perception~\cite{Zhang2024-wq, Zhang2020-fs}. Despite this, there is still some recent work focuses on enhancing the pose accuracy of roadside perception using various sensor types. For instance, \cite{Yang2023-zy} introduces BEVHeight, an approach that predicts height from the ground rather than pixel-wise depth, facilitating distance-independent perception for camera-only systems. This work improved the accuracy of roadside perception for object detection and segmentation tasks, although it did not attempt to produce pose data to localize vehicles from roadside observations. In another study~\cite{Gu2021-gs}, Gu~\textit{et al.} employed reversed point cloud matching on an RSU to accurately estimate the pose of the ego-vehicle. The proposed method requires the infrastructure to acquire the ego-vehicle's point cloud, which is generated from the vehicle's computer aided-design (CAD) model. This approach is challenging to implement in real-world settings and could significantly increase V2I traffic data. The increase in data volume within the V2X network is problematic as it could affect safety-related and time-sensitive services, which should be avoided~\cite{Abboud2016-mx}.

\subsection{Vehicular Localization and Cooperative Localization}
In most of the state-of-the-art localization methods, a high-definition (HD) map is a critical tool for achieving centimeter-level localization for AVs~\cite{Chalvatzaras2022-wl}. However, maintaining HD maps can be extremely costly~\cite{Pannen2020-xo}, and localizing against outdated HD maps may compromise localization accuracy~\cite{Plachetka2020-qr}. Moreover, the features within HD maps are indispensable for achieving centimeter-level accuracy in localization~\cite{Javanmardi2018-us}. A lack of matchable features or temporary changes in the landscape could disrupt the localization algorithms, resulting in poor localization accuracy.

Similar to cooperative perception, which plays a crucial role in eliminating blind spots in vehicle sensors, cooperative localization significantly enhances vehicle localization accuracy, especially when environmental factors compromise the reliability of self-localization efforts. In a typical cooperative localization framework, the ego-vehicle receives data from either other connected vehicles~\cite{Yang2020-wz} or smart infrastructure~\cite{Gao2024-ot}. The latter, known as infrastructure-based cooperative localization, often results in superior accuracy due to the fixed and stable nature of infrastructure stations. Although radio wave-based methods have been explored for vehicle positioning, their utility in cooperative localization is constrained by the inherent inaccuracies in wave-based distance measurements~\cite{Ou2019-xv}. Andert~\textit{et al.}~\cite{Andert2022-uq} have developed a cooperative localization approach that integrates data from stationary infrastructure cameras with sensors onboard the ego-vehicle. Their findings indicate that integrating infrastructure-based sensors into the cooperative localization system leads to noticeable improvements in localization performance.

\section{Proposed Method}
Although map-based localization techniques generally exhibit greater robustness compared to GNSS-based methods, there are scenarios where the available map features are inadequate or unsuitable for effective feature matching, leading to reduced localization performance~\cite{Li2017-yr}. To address this limitation, we propose leveraging LiDAR-equipped infrastructure, already utilized for cooperative perception, to enhance the localization accuracy of CAVs.

\subsection{System Structure}
The proposed system is built on Autoware, an open-source software stack designed for self-driving vehicles, offering a comprehensive platform which includes localization as one of its core modules~\cite{TIER_IV_inc_undated-ur}. It uses LiDAR sensors for localization. By comparing the point cloud data collected in real-time from LiDAR sensors with environment with HD maps, Autoware can determine the vehicle's location efficiently. Techniques such as NDT are commonly employed for this purpose.

In the proposed method, smart infrastructure utilizes LiDAR sensors to detect the CAV's relative position, which is then converted to a global position using the coordinates of the RSU. According to the European standard~\cite{Etsi2019-kl}, a CAV can communicate with other intelligent transport systems stations (ITS-S) via CAMs. It is assumed that our infrastructure is capable of extracting information from CAMs regarding the ego-vehicle's dimensions.

The roadside localization process begins by capturing a reference LiDAR frame at various positions to facilitate background filtering. Subsequently, the RSU continuously compares the current frame with the reference frame to identify new points. As the ego-vehicle approaches the RSU, background points are filtered out, leaving only those points generated by the ego-vehicle. Once the RSU detects the ego-vehicle, it calculates the vehicle's estimated 2D pose. This position information is then transmitted back to the vehicle via V2I communications. The CAV receives the estimated pose from the RSU and combines it with the self-localization data from its onboard sensors to achieve enhanced localization accuracy. The detailed structure of this process is illustrated in Figure~\ref{fig:structure}.

\subsection{Improvement to Roadside LiDAR Perception}\label{sec:improvement}

\begin{figure}[hbt]
\centering
\includegraphics[width=\linewidth]{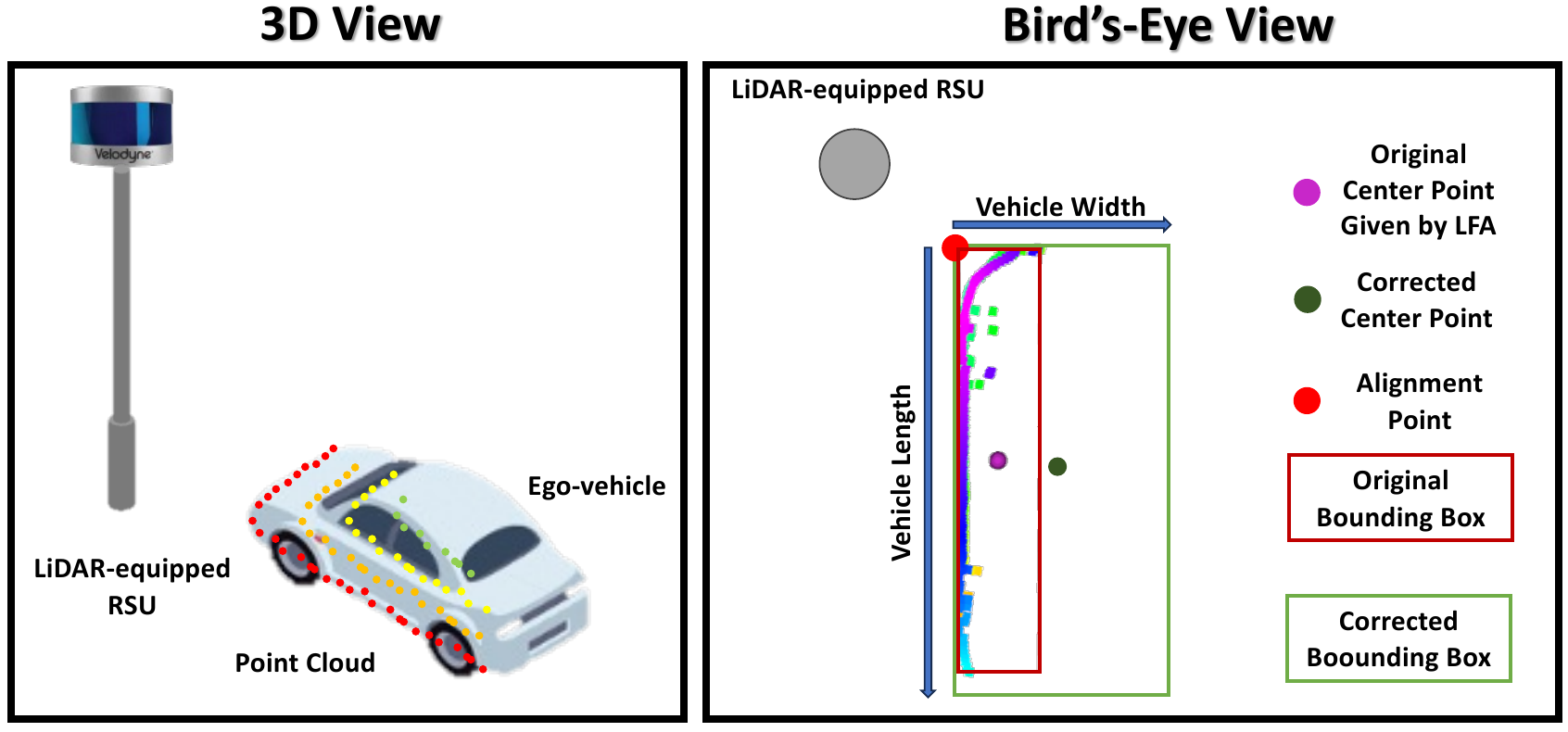}
\caption{Position refinement algorithm utilizing ground truth vehicle dimensions. Vehicle length and width data are transmitted from the CAV to the RSU via V2I communication. By aligning ground truth vehicle box to the alignment point, the RSU can estimate a more accurate vehicle pose.}
\label{fig:size_correction}
\end{figure}

Roadside LiDAR perception systems typically begin by performing background filtering, followed by clustering the remaining points into point clusters. Subsequently, an L-shape fitting algorithm (LFA) is employed to determine the bounding box of detected vehicles~\cite{Zhang2017-be}. In the original study, the authors focused their evaluation primarily on the heading error of the proposed algorithm, which demonstrated both accuracy and consistency. In our research, we assess positioning accuracy by measuring the distance between the center of the estimated bounding box and the vehicle's ground truth center position. Due to the vehicle's self-occlusion, the positioning accuracy is found to be less reliable than the heading angle accuracy~\cite{Jiang2023-gz}. This limitation underscores the need for further improvements in positioning accuracy for its application in cooperative localization.

Considering that a vehicle's dimensions remain constant over time, we propose leveraging V2I communication, which enables the RSU to acquire dimension information from the CAM of the connected vehicle, allowing for more accurate position estimation for localization purposes. With the ground truth dimensions of a vehicle's bounding box known, the RSU can precisely adjust the estimated bounding box by aligning it to one of the corner points. Among the four corner points determined by the LFA, the corner closest to the roadside LiDAR is identified as the most reliable for alignment. We designate this point as the `alignment point'. Aligning the dimension box with this corner allows for refined adjustment of the vehicle’s position, thereby enhancing localization accuracy. Figure~\ref{fig:size_correction} illustrates an example of the position estimation algorithm.

Although our method estimates the 2D pose of the vehicle, we also utilize the 3D information available from the point cloud. Typically, the width and length measurements pertain to the vehicle's chassis, which means that these dimensions are most accurate at the lower part of the vehicle. The upper parts of the vehicle, having smaller dimensions, do not contribute as accurately to pose estimation in our method. Moreover, we have identified that external features such as rear-view mirrors can adversely affect the performance of the LFA. Since LFA requires that all sensed points be contained within the bounding box, the inclusion of points from the rear-view mirrors can skew the estimated bounding box, leading to erroneous results. To address this issue, we propose using only those points in the LFA where the height is less than 0.8 meters, aiming for a more accurate and efficient bounding box estimation. This height threshold could potentially be adjusted on a vehicle-to-vehicle basis to optimize accuracy.

\subsection{Data Fusion}
In our experiment, we employed Autoware as the primary platform for autonomous driving and data fusion. All vehicle perception inputs, including LiDAR, GNSS, and IMU, were simulated and provided by AWSIM.

Autoware employs an Extended Kalman Filter (EKF) to integrate and smooth noisy sensor data into a cohesive position estimate. The implementation of EKF in Autoware includes the following major features:
\begin{itemize}
    \item \textbf{Time Delay Compensation}: This feature ensures accurate synchronization and integration of sensor data with varying time delays, which is critical for the performance and safety of high-speed autonomous vehicles.
    \item \textbf{Smooth Update Mechanism}: By distributing the impact of new measurements over multiple updates, this mechanism provides a smoother and more consistent localization. This is particularly beneficial for handling low-frequency data, enhancing the stability of the position estimates.
\end{itemize}

In our experiments, we employed the EKF module from Autoware for data fusion purposes. The output from the EKF module serves two primary functions: it is used as the definitive localization result for the AV, and it also provides the initial guess for the NDT scan matcher. The inputs to the EKF module include the estimated pose and the corresponding covariance error matrix from each data source's sensors. Given that Autoware's localization system operates with 6 degrees of freedom (DoF), it necessitates the use of the following covariance matrix:
\begin{equation}
\Sigma(\xi) =
\left[
\begin{smallmatrix}
\sigma_x & 0 & 0 & 0 & 0 & 0 \\
0 & \sigma_y & 0 & 0 & 0 & 0 \\
0 & 0 & \sigma_z & 0 & 0 & 0 \\
0 & 0 & 0 & \sigma_\phi & 0 & 0 \\
0 & 0 & 0 & 0 & \sigma_\theta & 0 \\
0 & 0 & 0 & 0 & 0 & \sigma_\psi \\
\end{smallmatrix} 
\right]
\end{equation}
where \(\xi = [\sigma_x, \sigma_y, \sigma_z, \sigma_\phi, \sigma_\theta, \sigma_\psi]^\intercal\) represents the standard deviations of the respective variables. Given that our RSU is a 2D localizer, we specifically configured its covariance matrix to focus on the horizontal plane, ignoring vertical and rotational uncertainties:

\begin{equation}
\xi_{\textrm{RSU}} = [\sigma_x^{\textrm{RSU}}, \sigma_y^{\textrm{RSU}}, \infty, \infty, \infty, \infty]^\intercal
\end{equation}
Standard deviations for the Velodyne LiDAR models used in our system are set as follows:

\begin{equation}
\sigma_x^{\textrm{VLP-16}} = \sigma_y^{\textrm{VLP-16}} = 0.01486
\end{equation}

\begin{equation}
\sigma_x^{\textrm{VLP-32C}} = \sigma_y^{\textrm{VLP-32C}} = 0.00681
\end{equation}

The covariance matrix for the NDT scan matcher is configured as follows:

\begin{equation}
\xi_\textrm{NDT} = [0.0225, 0.0225, 0.0225, 0.000625, 0.000625, 0.000625]^\intercal
\end{equation}

\subsection{Performance Optimization}

Localization for fast-moving robots, such as vehicles, is critically time-sensitive~\cite{Kuutti2018-rs}. In the context of a cooperative localization scheme, end-to-end delay primarily arises from two sources: data processing time and network delay. To address network delays, we will evaluate our method under a variety of network conditions. The findings and discussions pertaining to network related evaluation will be detailed in Section~\ref{sec:network_delay}. In this section, we will discuss the performance optimization we used to reduce data processing time.

\begin{figure}[htbp]
\centering
\includegraphics[width=\linewidth]{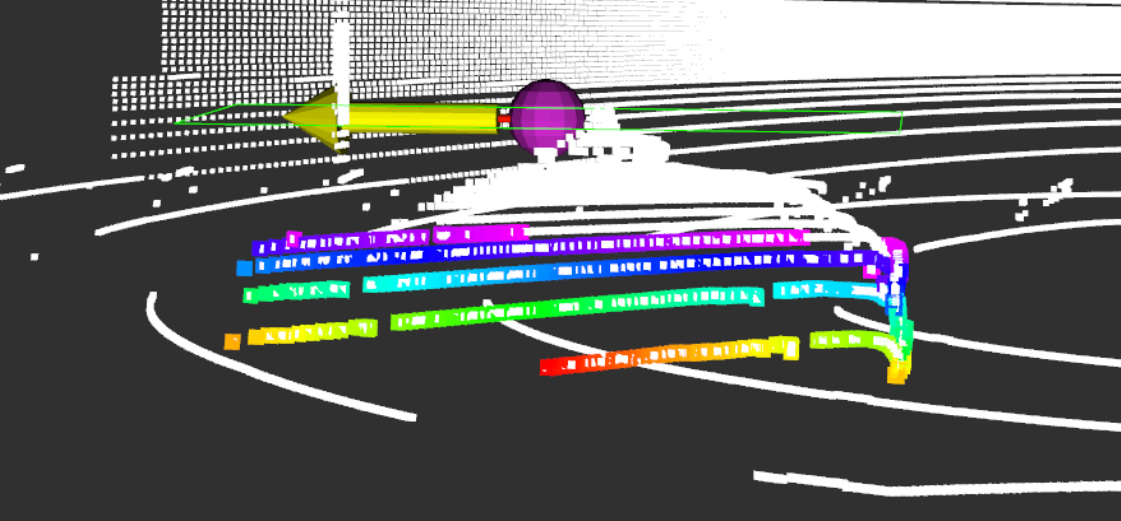}
\caption{This figure illustrates the point filtering process used in our position estimation algorithm. Points are sorted by height, and only the lower ones are retained for further analysis. In the visualization, colored points represent those used in the algorithm, while white points—including both background points and higher vehicle points—are filtered out.}
\label{fig:lower}
\end{figure}

In the data processing pipeline, background filtering and the LFA are the primary consumers of computation time. To efficiently distinguish vehicle points from background points, we employed a K-D tree to identify corresponding points between the current and reference frames. To further enhance the efficiency of background filtering, we leverage the reference position of the ego-vehicle. By restricting background filtering operations to the vicinity of this reference position and assuming other points are all background points, we significantly improve the performance of this process.

The LFA operates with a computational complexity of $O(N)$ with a very large constant factor, where $N$ represents the number of input points. When employing a high-resolution LiDAR and the vehicle is close to the roadside LiDAR sensor, there is a rapid increase in the number of points collected. To manage computation time effectively, we propose setting a maximum threshold for the number of input points. Since the lower points typically preserve the vehicle's contour adequately, we sort all collected points by height and retain only the bottom 500 points as the input point cloud for LFA. Figure~\ref{fig:lower} shows the point filtering results of a point cloud frame, where only the colored points are used in the LFA. 

\section{Experiment}

\begin{figure}[htb]
    \centering
    \includegraphics[width=0.7\linewidth]{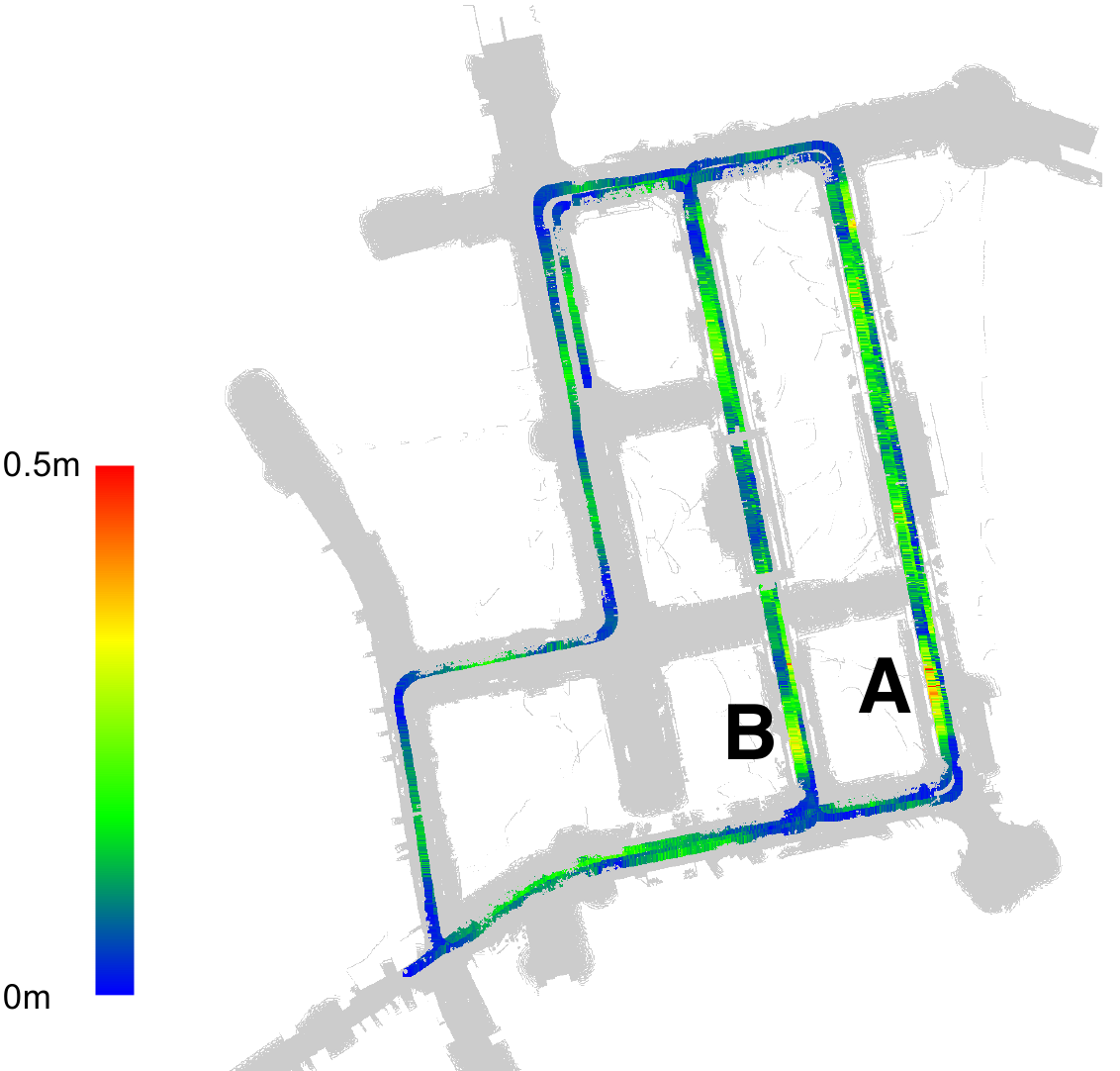}
    \caption{Illustration of the localization error evaluation conducted on the Nishishinkjuku map~\cite{Matsumoto2023-wr}. Locations A and B were selected as the experimental sites for our study.}
    \label{fig:matsumoto}
\end{figure}

To assess the performance of our proposed method under realistic conditions, we conducted tests using AWSIM, a digital twin simulator specifically designed for autonomous driving applications. The environment used in the simulator is based on Nishishinjuku, Tokyo, and its point cloud map is served as the simulation test bed. Drawing on findings from a previous study~\cite{Matsumoto2023-wr}, which identified locations A and B (as shown in Figure~\ref{fig:matsumoto}) as having the highest localization errors with NDT scan matching-based localization, we strategically placed roadside LiDAR sensors at these locations. These sensors are utilized to enhance vehicle self-localization capabilities through V2I communication and data fusion with onboard NDT-based localization systems. We assume that the dedicated short-range communications (DSRC) is used for the V2I communication.

In our experiments, each roadside LiDAR was mounted at a height of 2 meters. We employed two types of LiDAR models, the VLP-16 and VLP-32C, to evaluate their performance characteristics. According to a previous study~\cite{Jiang2023-gz}, the effective ranges for the VLP-16 and VLP-32C are 30 meters and 50 meters, respectively. For each trial, the vehicle was set to start from the same point and directed towards a predetermined destination. Upon activation of the Autoware autonomous driving mode, the vehicle proceeded past the LiDAR sensors. When the vehicle entered within the effective range of a RSU, the RSU began estimating the vehicle's pose at a frequency of 10 Hz. These pose estimations were then transmitted to the Autoware system via ROS messages to facilitate further data fusion. We developed the RSU localization module as a ROS 2 node, which is configured to subscribe to the output from the roadside LiDAR sensors. This node processes the data to determine localization results, which are then published to the vehicle.

\section{Results and Evaluation}
\subsection{Localization Performance}

\begin{figure}[htb]
    \centering
    \subfloat[Scenario A] {
        \includegraphics[width=0.4\textwidth]{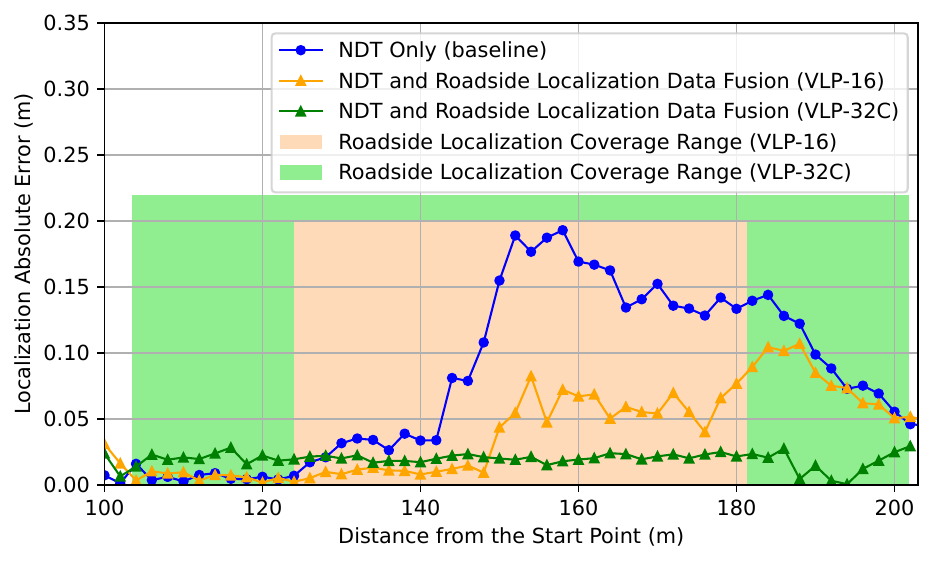}
        \label{fig:error_a}
    }\hfill
    \subfloat[Scenario B] {
        \includegraphics[width=0.4\textwidth]{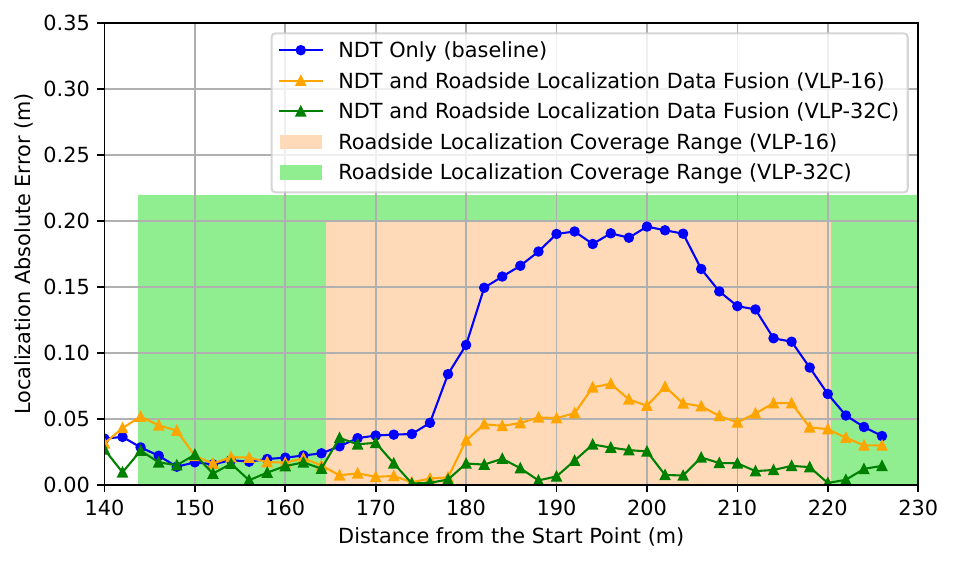}
        \label{fig:error_b}
    }
    \caption{Comparison of localization errors between the baseline and proposed methods using two different LiDAR models.}
    \label{fig:error_ab}
\end{figure}

We conducted experiments in two scenarios (A and B), both with and without our proposed method. During the trials, we used two different LiDAR models, VLP-16 and VLP-32C, as roadside sensors. To assess localization accuracy, we measured the minimum localization error each time the vehicle passed every 2 meters along the road. The results are documented in Figure~\ref{fig:error_ab}.

The data clearly show that within the coverage ranges of the deployed LiDARs, our proposed method significantly reduces localization errors. Specifically, while the error for NDT-only localization reached up to 20 cm in both scenarios, the error with our method did not exceed 10 cm within the coverage areas. Notably, when using the VLP-32C, the localization error consistently remained below 5 cm.

However, using a more powerful LiDAR model did not always yield better performance. For instance, between 100 m and 150 m as shown in Figure~\ref{fig:error_a}, the results from the VLP-16 (the less powerful model) actually outperformed those from the VLP-32C. We hypothesize that the increased number of points processed by the VLP-32C might lead to computational delays, potentially contributing to higher errors. Despite this, the overall performance of the VLP-32C was still superior.

Note that in Scenario A, around the 150 m mark, the cooperative localization error increased when using the VLP-16. This increase is attributed to the inaccuracy in the vehicle's self-localization, which relies on NDT scan matching. Since our method involves the data fusion of self-localization and the estimated pose from the RSU, any inaccuracy in one data source can compromise the overall accuracy. Conversely, the VLP-32C, which utilizes a smaller variance value, maintained accurate and stable performance in cooperative localization.

\begin{figure*}[t]
    \centering
    \subfloat {
        \centering
        \includegraphics[width=0.32\textwidth]{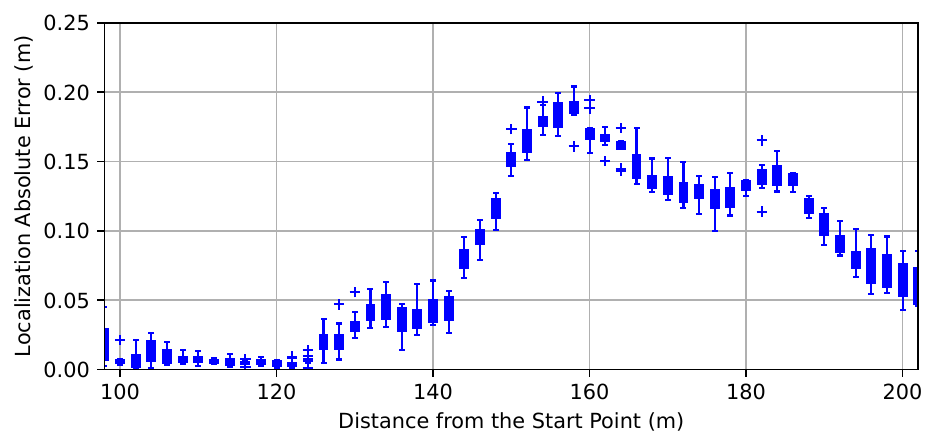}
    }
    \subfloat {
        \centering
        \includegraphics[width=0.32\textwidth]{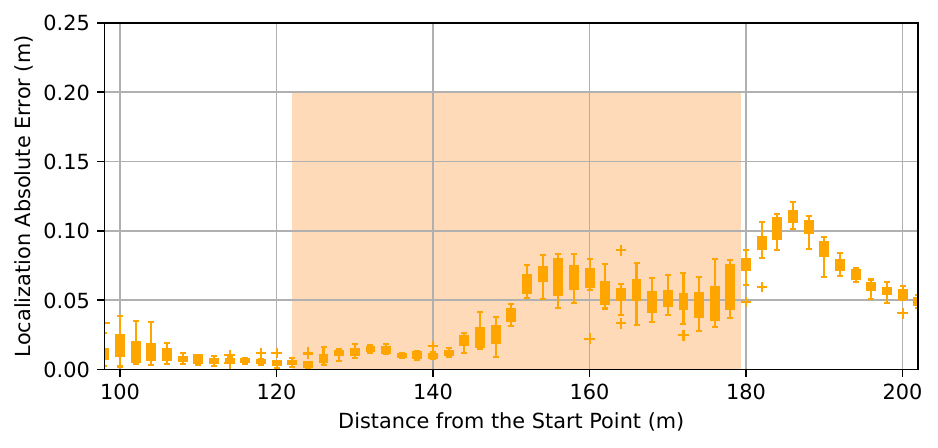}
    }
    \subfloat {
        \centering
        \includegraphics[width=0.32\textwidth]{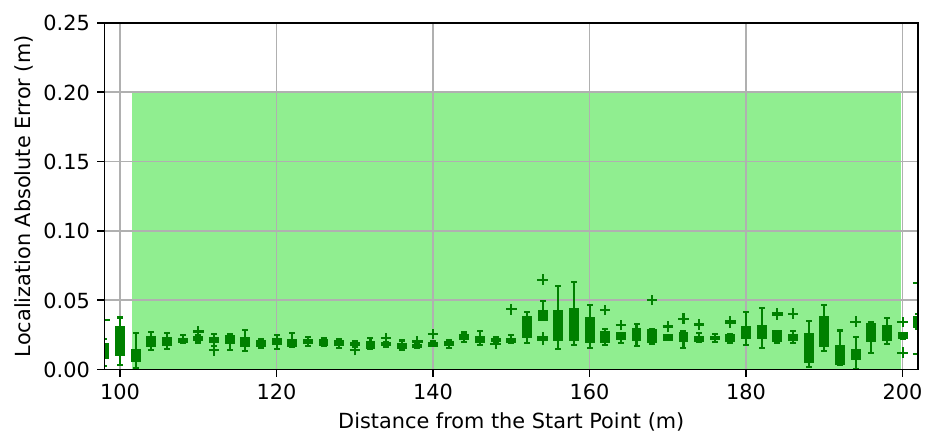}
    }\hfill
    \subfloat {
        \centering
        \includegraphics[width=0.32\textwidth]{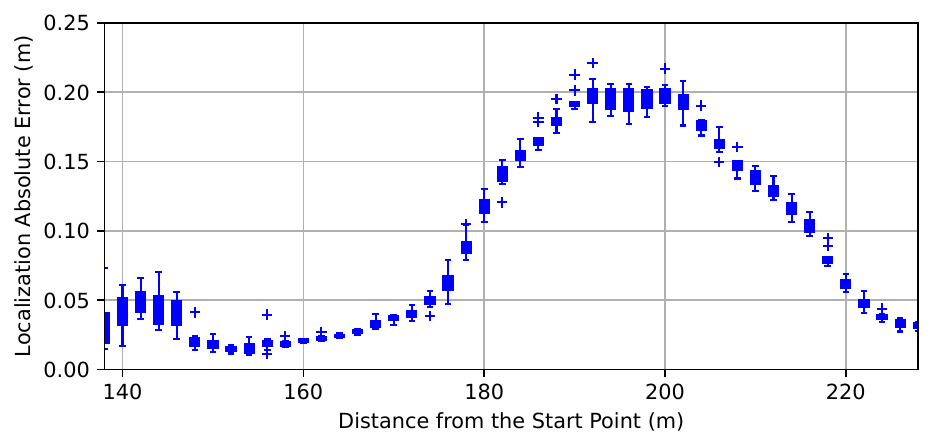}
    }
    \subfloat {
        \centering
        \includegraphics[width=0.32\textwidth]{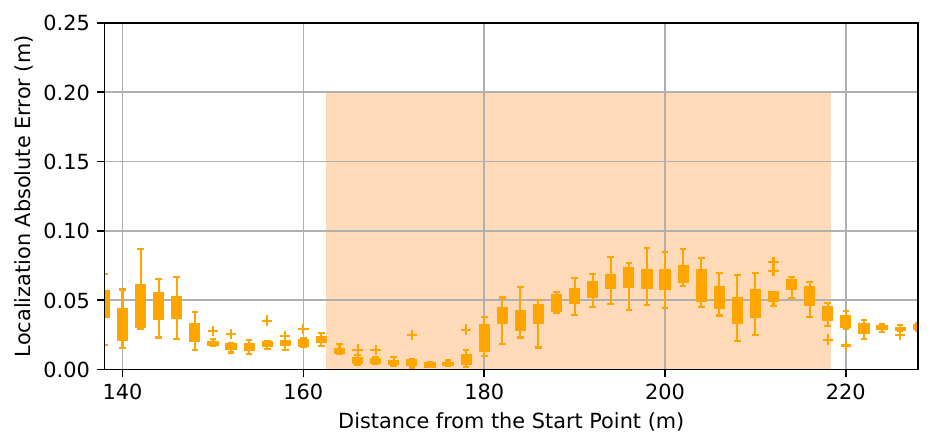}
    }
    \subfloat {
        \centering
        \includegraphics[width=0.32\textwidth]{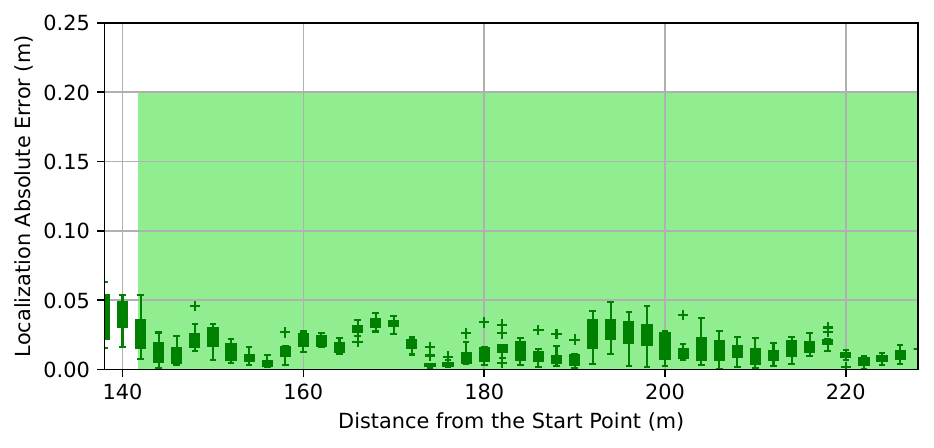}
    }
    \caption{Localization performance comparison across two scenarios over 10 trials. The figures in the first row (Scenario A) and the second row (Scenario B) illustrate the localization results. The blue plots represent the baseline NDT-only localization. The orange plots depict cooperative localization using the VLP-16 roadside LiDAR, while the green plots indicate cooperative localization with the VLP-32C roadside LiDAR. The respective color boxes in each plot demarcate the coverage range of the roadside localization for each LiDAR model.}
    \label{fig:10_ab}
\end{figure*}

To verify the consistency of our proposed method, we repeated the experiment in each scenario 10 times. The collective results are presented in Figure~\ref{fig:10_ab}. On average, our method significantly enhanced the localization accuracy compared to NDT scan matching. Specifically, when using the VLP-16 as the roadside LiDAR, the localization error improved by 64\% to 69\%. With the VLP-32C, the improvement ranged from 76\% to 83\%. Detailed statistical data are available in Table~\ref{tab:mean_error}.

\begin{table}[htb]
\centering
\caption{Mean Localization Error (MLE) (m)}
\begin{tabular}{@{}lll@{}}
\toprule
Methods                     & Scenario A & Scenario B \\ \midrule
NDT only in VLP-16 range (baseline)   & 0.0984     & 0.1265     \\
NDT \& VLP-16 Fusion        & 0.0353 \textcolor{green}{$\downarrow$64\%}    & 0.0391     \textcolor{green}{$\downarrow$69\%} \\
NDT only in VLP-32C range (baseline) & 0.0827     & 0.0928     \\
NDT \& VLP-32C Fusion       & 0.0224 \textcolor{green}{$\downarrow$76\%}    & 0.0157   \textcolor{green}{$\downarrow$83\%}  \\ \bottomrule
\end{tabular}
\label{tab:mean_error}
\end{table}

\subsection{Resistance Against Network Delay and Packet Loss}\label{sec:network_delay}

\begin{figure}[htbp]
    \centering
    \includegraphics[width=\linewidth]{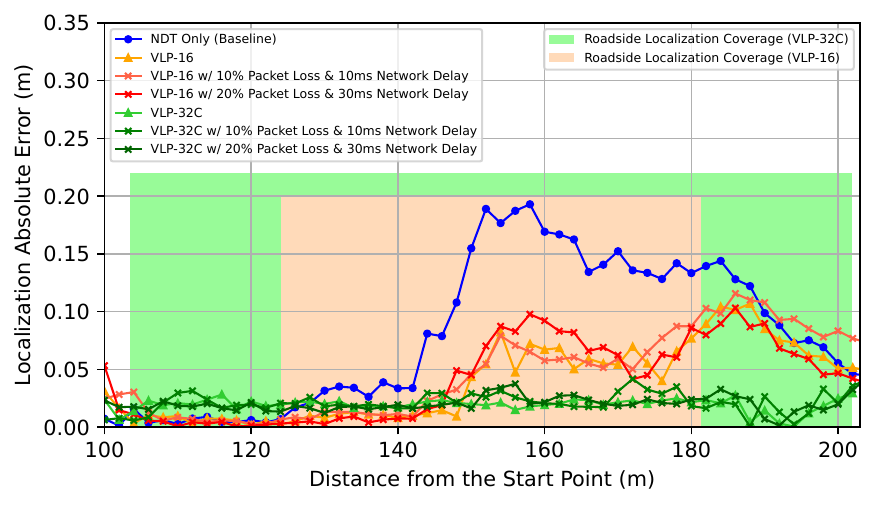}
    \caption{Evaluating the performance of the proposed method under adverse network conditions. The experiments depicted in this figure were conducted in scenario A.}
    \label{fig:network_simulation}
\end{figure}

Due to the non-instantaneous nature of sensing, communication, and data processing, there is often a discrepancy between the time an object is sensed and when the localization result becomes available. Hence, it is crucial to account for network delays and packet loss when evaluating our proposed method. Following the findings from studies on packet loss~\cite{Jiang2010-ep} and network delay~\cite{Asabe2023-jk} in V2X networks, we configured our system to simulate usual network conditions with a 10 ms delay and 10\% packet loss, and poor network conditions with a 30 ms delay and 20\% packet loss. The impact of these conditions on localization accuracy is illustrated in Figure~\ref{fig:network_simulation}. Under poor network conditions, an increase in localization error was observed. Nevertheless, our proposed method still effectively reduced the localization error compared to the baseline.

\begin{table}[htbp]
\centering
\caption{Localization error in different network conditions (VLP-16)}
\begin{tabular}{|l|lll|}
\hline
\multirow{1}{*}{MLE (m)} & \multicolumn{3}{c|}{Network Delay}                            \\ \cline{1-4} 
Packet Loss& \multicolumn{1}{l|}{0 ms}      & \multicolumn{1}{l|}{10 ms}     & 30 ms    \\ \hline
0\% & \multicolumn{1}{l|}{0.0316} & \multicolumn{1}{l|}{0.0317 \textcolor{red}{$\downarrow$0.3\%}} & 0.0318 \textcolor{red}{$\downarrow$0.6\%}\\ \hline
10\% & \multicolumn{1}{l|}{0.0306 \textcolor{green}{$\uparrow$0.3\%}} & \multicolumn{1}{l|}{0.0327 \textcolor{red}{$\downarrow$3.5\%}} & 0.0337 \textcolor{red}{$\downarrow$6.6\%}\\ \hline
20\% & \multicolumn{1}{l|}{0.0333 \textcolor{red}{$\downarrow$5.4\%} } & \multicolumn{1}{l|}{0.0341 \textcolor{red}{$\downarrow$7.9\%} } & 0.0346 \textcolor{red}{$\downarrow$9.5\%} \\ \hline
\end{tabular}
\label{tab:vlp16_network}
\end{table}

\begin{table}[htbp]
\centering
\caption{Localization error in different network conditions (VLP-32C)}
\begin{tabular}{|l|lll|}
\hline
\multirow{1}{*}{MLE (m)} & \multicolumn{3}{c|}{Network Delay}                            \\ \cline{1-4} 
Packet Loss& \multicolumn{1}{l|}{0 ms}      & \multicolumn{1}{l|}{10 ms}     & 30 ms     \\ \hline
0\% & \multicolumn{1}{l|}{0.0208} & \multicolumn{1}{l|}{0.0224 \textcolor{red}{$\downarrow$7.7\%}} & 0.0233 \textcolor{red}{$\downarrow$12.0\%} \\ \hline
10\% & \multicolumn{1}{l|}{0.0239 \textcolor{red}{$\downarrow$14.9\%}} & \multicolumn{1}{l|}{0.0226 \textcolor{red}{$\downarrow$8.7\%}} & 0.0248 \textcolor{red}{$\downarrow$19.2\%} \\ \hline
20\% & \multicolumn{1}{l|}{0.0223 \textcolor{red}{$\downarrow$7.2\%}} & \multicolumn{1}{l|}{0.0225 \textcolor{red}{$\downarrow$8.2\%}} & 0.0223 \textcolor{red}{$\downarrow$7.2\%} \\ \hline
\end{tabular}
\label{tab:vlp32c_network}
\end{table}

In Table~\ref{tab:vlp16_network} and Table~\ref{tab:vlp32c_network}, we present the average localization errors for the VLP-16 and VLP-32C LiDARs, respectively, across their effective ranges under various network conditions. The VLP-16 exhibits an approximate 10\% decrease in accuracy in the worst-case network scenarios, while the VLP-32C shows a more pronounced accuracy reduction of about 20\%. Notably, the VLP-16 demonstrates greater resilience to adverse network conditions.

\section{Conclusion}

In conclusion, this paper has explored the effectiveness of roadside LiDAR cooperative localization with the Autoware AD framework. First, we proposed a method to significantly enhance the accuracy of roadside LiDAR perception in terms of positioning error. Subsequently, we implemented a data fusion scheme that integrates on-board and roadside localization results. By conducting realistic simulations, we demonstrated how the inclusion of roadside LiDAR can enable more precise and reliable vehicle localization. The experimental data shows that our proposed method can improve localization accuracy by 60\% to 80\%. Besides, our method also demonstrated good resistance against various network conditions.

Our proposed method contributed to the advancement of autonomous vehicle technology by achieving the following key objectives. Firstly, it reduced localization errors compared to conventional NDT-based vehicle self localization, particularly in vulnerable areas, thereby enhancing the overall safety of autonomous vehicles. Additionally, it leveraged smart infrastructure to provide cooperative localization services, thereby increasing the infrastructure's utility and value by enabling the provision of additional services.

It is also important to highlight the scalability and adaptability of our method when applied to scenarios involving multiple roadside LiDARs. The deployment of additional LiDAR units will not only extend the geographical coverage but also significantly enhance the robustness and reliability of the roadside localization system by eliminating possible occlusions caused by vehicles and other infrastructure. This capability demonstrates the potential of our approach in more extensive and complex environments.

Looking ahead, our future work will focus on several areas to refine and expand upon our current findings. We aim to conduct real-world experiments that will integrate advanced roadside LiDAR background filtering techniques, such as 3D-DSF~\cite{Wu2017-ym}, to cope with complex environments. Additionally, we plan to address scenarios involving multiple vehicles, taking into account the challenges posed by occlusions. Finally, we will investigate methods for sensor fusion to effectively combine data from multiple roadside LiDARs for enhanced accuracy and reliability.

\section*{Acknowledgment}
This work was supported by JST ASPIRE Grant Number JPMJAP2325, Japan. Partial financial support was received from TIER IV, Inc.

\bibliographystyle{ieeetr}
\bibliography{main}

\end{document}